\documentclass[letterpaper, 10 pt, conference]{ieeeconf}  

\IEEEoverridecommandlockouts                              

\usepackage{amsmath} 
\usepackage{amssymb}  
\usepackage{graphicx}
\usepackage[table]{xcolor}

\title{\LARGE \bf
Evaluating Social Reasoning in Embodied Vision-Language Models
}

\author{Daniel Weiner$^{1}$ and Raj Korpan$^{2}$%
\thanks{$^{1}$Daniel Weiner is with The Graduate Center, City University of New York (CUNY), New York, NY, USA
        {\tt\small dweiner@gradcenter.cuny.edu}}%
\thanks{$^{2}$Raj Korpan is with Hunter College and The Graduate Center, City University of New York (CUNY), New York, NY, USA
        {\tt\small raj.korpan@hunter.cuny.edu}}%
}

\begin{document}

\maketitle
\thispagestyle{empty}
\pagestyle{empty}

\begin{abstract}
Foundation models now serve as reasoning layers for robot teams, but it remains unclear whether embodiment-specialized vision-language models (VLMs) preserve the social-reasoning capabilities of general-purpose VLMs. We evaluate Gemini Robotics-ER on SHREC, a benchmark of real human-robot social interactions, and compare it with published general-purpose VLM baselines. Robotics-ER outperforms GPT-4o-mini on error/competence detection but performs worse on detailed social-attribute judgments; adding a visual frame helps one task while hurting the other. These results suggest embodiment-specialized models may perform differently on social judgments that can be read more directly from observable interaction cues versus those requiring more inference about a person’s internal state. Because the comparison spans model families, these differences cannot be attributed to embodiment specialization alone. This matters for multi-robot systems, where a shared foundation model could cause systematic social-perception errors to influence multiple coordination decisions. This work presents an initial evaluation, with broader cross-context and video-based analyses left for future work.
\end{abstract}


\section{Introduction and Related Work}
\label{sec:introduction}

Large language models (LLMs) and vision-language models (VLMs) have quickly moved from perception and planning aids to the reasoning core of robot autonomy \cite{ichter2022saycan, zitkovich2023RT2}. 
A distinct class of \emph{embodiment-specialized} models for robotic reasoning has emerged, targeting spatial grounding, pointing, affordance and trajectory prediction, and success detection \cite{geminiroboticsteam2025geminiroboticsbringingai}. 
Such models are natural candidates for high-level reasoning about what a robot should do from multimodal observations. Increasingly, those decisions are made among \emph{people}, and by \emph{teams} of robots: guides in airports, assistants in hospitals, service fleets in malls.
For such teams, coordination quality hinges on a shared assessment of human social state, such as who is engaged, who is confused, who is disengaging, or whether a robot has made a social error, so the team can determine which robot should approach, continue an interaction, or withdraw. 
We refer to this as a \emph{social-perception layer}: a semantically rich estimate of the humans in view produced from each robot's observations and made available to the coordination layer.

Recent work has begun evaluating whether general-purpose VLMs can provide such social information. SHREC benchmarks foundation models on real human-robot interactions (HRI), including recognition of social errors, competencies, and social attributes such as engagement, and reports substantial variation across models and tasks \cite{lee2025social}. HRIBench similarly evaluates VLMs on interaction-relevant human perception, including nonverbal cues and other visual signals \cite{shi2025hribench}, while complementary work studies broader interaction-centric human-robot collaboration capabilities \cite{liu2026hribench}. VLMs have also been used directly as proxies for social intelligence \cite{bu2025using} and to determine whether and when a robot should initiate interaction \cite{sasabuchi2025agreeing}.

Embodiment-specialized models are appealing candidates for the social-perception layer: one pretrained model, no per-deployment training, outputs already expressed in language the rest of the robotics stack uses \cite{geminiroboticsteam2025geminiroboticsbringingai}. 
Existing HRI benchmarks primarily evaluate general-purpose VLMs, while embodiment-specialized models are primarily evaluated on physical and spatial capabilities. To our knowledge, the relative social-reasoning performance of an embodiment-specialized VLM has not been evaluated against general-purpose VLM baselines on the same HRI benchmark.

Concretely, we evaluate whether Gemini Robotics-ER, a model tuned for embodied robotic reasoning, can identify human social states, such as engagement, robot social competencies, and social errors, on the SHREC benchmark \cite{lee2026shrec}, and compare its performance with the general-purpose VLMs reported by SHREC. SHREC's social labels vary in how directly they can be supported by observable interaction evidence, thus, we examine performance along a gradient from externally grounded to internally inferred. Externally grounded tasks can be resolved from the observable structure of an interaction, such as turn-taking and response timing evident in the transcript. Internally inferred tasks require stronger inference about affect, intention, or mental state. Our hypothesis is that Gemini Robotics-ER will perform better on externally grounded tasks than internally inferred ones.


Furthermore, foundation models are increasingly being placed directly inside multi-robot decision-making loops. RoCo uses language-model agents to negotiate collaborative plans \cite{mandi2024roco}; SMART-LLM decomposes natural-language tasks and forms robot coalitions based on robot capabilities \cite{kannan2024smart}; and COHERENT coordinates heterogeneous teams through LLM-mediated task assignment and feedback \cite{liu2025coherent}. 
Recent work applies conformal prediction and uncertainty-aware reasoning to robot planning \cite{wang2024probabilistically,liang2024introspective, luo2024sample}, while embodied-agent evaluations show that models may fail to abstain appropriately under ambiguous or infeasible conditions \cite{yeke2026yes}. Recent work also argues for deployment-specific evaluation when selecting foundation models for social robots \cite{nichols2026should}. Against this background, we ask whether an embodiment-specialized robotics VLM preserves the social-reasoning capabilities needed for human-aware robotic systems.

We make three contributions. First, we formulate the question of whether an embodiment-specialized robotics foundation model preserves social reasoning capability, particularly in the context of human-aware multi-robot systems. Second, we define a benchmark-aligned evaluation of Gemini Robotics-ER on SHREC, validating our modified evaluation pipeline against an established general-purpose VLM baseline. Third, we analyze the implications of any observed capability gap for model reliability across multi-robot teams.

\section{Problem Formulation and Methodology}
\label{sec:method}

\subsection{Problem Setup and Dataset}
Consider a heterogeneous robot team operating in a human-populated environment. 
Each robot $r$ observes nearby people from its own viewpoint and produces a social percept $s_r$ that it provides to a shared coordination layer that allocates human-facing tasks, such as approach, assist, hand off, or disengage. 
We treat social perception as a mapping from an egocentric observation, consisting of sampled visual frame(s) plus available dialogue, to a structured assessment of engagement, social errors, or competencies, along with a confidence score. Model outputs are represented in a typed, machine-readable schema so that predictions can be used by downstream coordination modules.

Let $M_{\text{emb}}$ be an embodiment-specialized model (e.g., Gemini Robotics-ER) and $\{M_{\text{gen}}\}$ the set of general-purpose VLMs benchmarked on SHREC. For a social-reasoning task $T$ with metric $\mu_T$, we compare $\mu_T(M_{\text{emb}})$ with the corresponding performance of $\mu_T(\{M_{\text{gen}}\})$ across engagement, competence, and error detection.
Our primary hypothesis is that the relative performance of $M_{\text{emb}}$ will depend on the type of inference required by the task: embodiment-specialized models will perform better on externally grounded tasks than on internally inferred ones. If embodiment-specialized models perform better on both types of tasks compared to general-purpose VLMs, then that would indicate that embodiment specialization broadly preserves social reasoning, whereas if it were worse for both task types, it would indicate a genuine tradeoff between embodiment specialization and social reasoning.

The SHREC dataset \cite{lee2026shrec} comprises real human-robot interactions drawn from month-long, in-the-wild deployments of social robots across three deployment contexts: empathic storytelling in homes, wellness coaching in dormitories, and wellness coaching in homes. 
Each sample provides a sequence of extracted still frames, a dialogue transcript, and segment-level annotations from three independent human annotators.
Raw video is available only for the wellness-home subset at the time of writing. 
The empathic and wellness-dorm subsets provide still frames and transcripts without video.
Annotations mark whether the robot committed a social \emph{error} or exhibited a \emph{competence} per segment, together with seven social attribute flags (including \emph{engagement}), a rationale, and a suggested correction. 
We use the empathic subset for the primary quantitative evaluation. We additionally use a small sample from wellness-home for an exploratory transcript-versus-frame-versus-video comparison because it is the only subset for which raw video is currently available.

\subsection{Evaluation Setup}
We evaluate the two tasks for which SHREC reports general-model baselines, following the benchmark's task definitions and scoring procedure: (i) social error/competence detection, scored as classification among error, competence, and none; and (ii) social-attribute identification, including engagement, scored using SHREC's partial-match metric.

We evaluate Gemini Robotics-ER (v2.0), an embodiment-specialized reasoning model accessed via API, in a zero-shot setting with decoding fixed to temperature $0$ for reproducibility.
Outputs are constrained to a typed schema so predictions parse deterministically into the task label space.
A matched within-family comparison against SHREC's reported Gemini baselines was not possible because the corresponding Gemini endpoints are no longer available. We therefore use GPT-4o-mini, for which SHREC reports published results, as the reproduced general-purpose comparison model.
The primary comparison is therefore between an embodiment-specialized Gemini model and a general-purpose GPT model. Because model-family differences are confounded with specialization, we interpret the results as a comparison of these deployed model configurations rather than as a causal estimate of the effect of embodiment specialization. This limitation is discussed in Section~\ref{sec:discussion}.

We use three input conditions throughout the evaluation: transcript only (\textbf{T}), transcript plus a single still frame (\textbf{T+F}), and transcript plus video (\textbf{T+V}). The empathic subset does not include raw video, so its visual input is drawn from the publicly released sequence of still frames. Accordingly, Gemini Robotics-ER is evaluated under T and T+F on the empathic subset. SHREC's published GPT-4o-mini results provide T and T+V conditions.
The T condition is the directly matched input condition across Gemini Robotics-ER and the published GPT-4o-mini baseline and therefore serves as our primary model-to-model comparison. SHREC reports little or no improvement from adding visual input on these tasks, but we additionally evaluate Robotics-ER under T+F to measure the marginal effect of adding a still frame. Because T+F and T+V provide different visual information, we report these multimodal conditions for context but do not treat them as a controlled model-to-model comparison.

On the wellness-home subset, where raw video is available, we additionally conduct a small exploratory modality probe using all three conditions: T, T+F, and T+V. 
The probe contains five examples and is descriptive rather than quantitative; its purpose is to examine whether adding still-image or temporal visual information changes Robotics-ER's predicted social assessment or stated reasoning.

\subsection{Scoring and Validation}
The comparison is only meaningful if our evaluation pipeline scores predictions compatibly with SHREC's published procedure. Because SHREC's released evaluation harness was not designed to interface with Gemini Robotics-ER, we modified the model-invocation and output-parsing components so that Gemini predictions could be evaluated within the same pipeline, while leaving the benchmark's scoring logic unchanged.
We therefore first run GPT-4o-mini through our evaluation pipeline on a 25-sample subset of the language-only detection task and compare the results with SHREC's published GPT-4o-mini baseline. Our score falls within $0.06$ of the published value; across repeated runs, the reproduced baseline has a standard deviation of $\pm 0.02$. We treat this as a scoring-consistency check showing that the modified evaluation harness produces values in the neighborhood of the published evaluation, rather than as a full reproduction of the benchmark result.

SHREC provides annotations from three independent annotators, whose labels may disagree. Because the handling of these disagreements directly affects the scored targets, we retain SHREC's aggregation rule: only labels with strict pairwise agreement are retained, and disagreements are discarded. This preserves the same ground-truth approach used for the published baselines.

\section{Preliminary Results}
\label{sec:preliminaryresults}


\begin{table}[t]
\centering
\caption{SHREC task accuracy. GPT-4o-mini values are published SHREC baselines; Gemini Robotics-ER values are from our evaluation. Attribute identification uses SHREC's partial-match metric.}
\label{tab:results}
\begin{tabular}{cccc}
Model              & Condition & Detection & Attribute ID \\ \hline
GPT-4o-mini         & T         & $0.26\pm0.02$ & $0.41\pm0.04$ \\
GPT-4o-mini         & T+V       & $0.20\pm0.03$ & $0.42\pm0.02$ \\ \hline
Gemini Robotics-ER & T         & $0.296$        & $0.382$ \\
Gemini Robotics-ER & T+F       & $0.311$        & $0.342$
\end{tabular}
\end{table}

\begin{table}[t]
\centering
\caption{Exploratory modality probe on five wellness-home examples. Highlighted cells indicate a prediction change across input conditions; (R) denotes a change in stated reasoning without a change in predicted label. Gold labels summarize the majority label across the interaction.}
\label{tab:probe}
\begin{tabular}{|c|c|c|c|c|}
\hline
Example & Gold (majority)        & T     & T+F   & T+V                           \\ \hline
P50.    & Comp. (3 err, 18 comp) & Comp. & Comp. & Comp.                         \\ \hline
P54     & Comp. (3 err, 14 comp) & Error & Error & Error                         \\ \hline
P57     & Tie (3 err, 3 comp)    & None  & None  & \cellcolor[HTML]{FFFE65}Comp. \\ \hline
P60     & Comp. (0 err, 1 comp)  & Comp. & Comp. & \cellcolor[HTML]{FFFE65}Error \\ \hline
P80     & Tie (17 err, 17 comp)  & Error & Comp. & Comp.(R)                      \\ \hline
\end{tabular}
\end{table}

Table~\ref{tab:results} reports preliminary results on the two SHREC tasks for which published GPT-4o-mini baselines are available: error/competence detection and social-attribute identification. As defined in Section~\ref{sec:method}, the T condition is the directly matched input condition across models and therefore provides the primary model-to-model comparison. The multimodal conditions are reported separately because Gemini Robotics-ER is evaluated with T+F, whereas SHREC reports GPT-4o-mini with T+V.

On error/competence \emph{detection}, Gemini Robotics-ER scores $0.296$ under T compared with the published GPT-4o-mini baseline of $0.26$, a difference of $0.036$. For Gemini Robotic-ER, accuracy increased slightly under T+F to $0.311$. For reference, SHREC reports $0.20$ under T+V for GPT-4o-mini; because this condition uses video rather than a single frame, we report it for context rather than as a directly matched comparison with T+F.
On the finer-grained \emph{social-attribute identification} task, the direction reverses: Gemini Robotics-ER scores $0.382$ under T, compared with the published GPT-4o-mini baseline of $0.41$, a difference of $-0.028$. For Gemini Robotics-ER, accuracy reduced under T+F to $0.342$. SHREC's published GPT-4o-mini accuracy is $0.42$ under T+V, again reported for context rather than as a directly matched comparison with T+F. These preliminary results show that on the matched T condition, Gemini Robotics-ER outperforms GPT-4o-mini on detection but underperforms it on attribute identification.

We next examine whether richer visual input qualitatively changes Gemini Robotics-ER's judgments on five examples from the wellness-home subset, where raw video is available.
As shown in Table~\ref{tab:probe}, the three input conditions do not always produce the same prediction.
Adding a single frame changed the model's predicted label in one case (P80). Adding video changed the predicted label in two cases (P57 and P60); in P80, the predicted label remained unchanged under T+V, but the model's stated reasoning changed.
The effect of video is not uniformly beneficial. For P57, video changes the prediction from \emph{None} under T and T+F to \emph{Competence} under T+V. For P60, the prediction changes from \emph{Competence} under T and T+F to \emph{Error} under T+V.
The gold column in Table~\ref{tab:probe} summarizes the distribution of annotated error and competence labels across the full interaction rather than serving as the segment-level scoring target used in the primary quantitative evaluation. Accordingly, this probe is descriptive and is not used to draw accuracy conclusions.

\section{Discussion}
\label{sec:discussion}
The preliminary results show different relative performance patterns across the two SHREC tasks. 
On error/competence detection, Gemini Robotics-ER scores $29.6\%$ under the directly matched transcript-only condition, compared with the published GPT-4o-mini baseline of $26\pm2\%$. 
Adding a single frame increased Gemini Robotics-ER accuracy to $31.1\%$, a $1.5\%$ improvement over transcript-only. This modest within-model gain is consistent with SHREC's broader observation that visual input provides limited additional benefit on the detection task.
Among the published SHREC detection baselines, Robotics-ER's $29.6\%$ transcript-only result would rank highest among the language-only conditions, while its $31.1\%$ T+F result would rank second overall, behind only Gemini-1.5-Flash at $32\pm1\%$ under T+V. However, these conditions are not directly matched since Robotics-ER receives a single still frame whereas Gemini-1.5-Flash receives video, and the deprecated Gemini endpoint prevents us from rerunning that model through our modified harness. We therefore treat this comparison as contextual rather than evidence of statistical equivalence.

The detection scores should also be interpreted in relation to the task's class structure. The \emph{None} class consists of unannotated conversational segments that are count-matched to annotated segments and represents a large fraction of the evaluation data. In our outputs, both models show a tendency to predict \emph{Social Competence}, which reduces accuracy on \emph{None} examples. This pattern suggests that distinguishing an unremarkable interaction from a socially competent one is an important source of error, particularly when only transcript information is available.

On social-attribute identification, Gemini Robotics-ER scores $38.2\%$ under T compared with the published GPT-4o-mini result of $41\pm4\%$, while adding a frame reduces Gemini Robotics-ER accuracy to $34.2\%$. 
Thus, on the directly matched transcript-only condition, Robotics-ER performs better than GPT-4o-mini on detection but worse on attribute identification. The results therefore do not show a uniform advantage or deficit for the embodiment-specialized model. Instead, the task-dependent pattern is consistent with the possibility of a specialization tradeoff. Because the models belong to different base families, however, we cannot attribute this pattern causally to embodiment specialization itself. This asymmetry is broadly consistent with our proposed distinction between more externally grounded and more internally inferred social judgments: coarse error/competence detection may depend more heavily on observable interaction structure, whereas attribute identification can require finer judgments about affect, intention, or engagement. At present, this remains an interpretation rather than a demonstrated mechanism. A label-level analysis is needed to determine whether Robotics-ER's errors actually concentrate on the more internally inferred attributes.

The decrease from T to T+F on attribute identification also indicates that additional visual input is not uniformly beneficial for Robotics-ER. One possible interpretation is that embodiment specialization biases the model toward attributes that are more directly tied to observable interaction cues, such as engagement or conversational mechanics, at the expense of attributes that require stronger inference about intention or internal state. The addition of a single frame may reinforce this tendency by emphasizing immediately visible cues rather than correcting it. This interpretation remains speculative and requires a label-level error analysis. SHREC's partial-match metric further complicates the comparison because it scores only the model's highest-ranked predicted attribute; an ordering bias can therefore reduce accuracy even when the model identifies other correct attributes. A full multi-label evaluation, such as F1 over all predicted attributes, could distinguish between failures to identify the correct attributes and failures to rank them first.

The five-example wellness-home probe provides preliminary evidence that input modality can change Gemini Robotics-ER's social judgments.
Adding a single frame changes the predicted label in one case (P80), while adding video changes the predicted label in two cases (P57 and P60); in P80, the label remains unchanged between T+F and T+V but the model's stated reasoning changes. The direction of these changes is not consistent across examples. Because the sample is too small for quantitative inference and the interaction-level gold summaries do not serve as the segment-level targets used in the primary evaluation, we do not interpret these changes as improvements or degradations in accuracy. They instead suggest that richer visual context can alter both the model's predicted social label and its accompanying reasoning.
This motivates a larger controlled video evaluation.

For multi-robot systems, the task-dependent performance observed here matters because social perception may become a shared input to task allocation and coordination.
A disengagement assessment, for example, could lower the suitability of a ``continue-interaction'' action and increase the suitability of a ``hand-off'' or ``disengage'' action in a task allocation method such as HVBTA \cite{weiner2025hybridvoting, weiner2025hybrid}. If the same foundation model is deployed across a fleet, systematic social-perception errors may propagate across robots rather than being averaged away through independent models. Reliable uncertainty estimates could eventually allow a coordination layer to abstain, choose a conservative action, or defer to another robot with a clearer view when a social assessment is uncertain. These downstream effects are motivating applications of the present study rather than outcomes measured here. 


These results have several limitations that also motivate future work. Because our primary comparison is between Gemini Robotics-ER and GPT-4o-mini, model-family and specialization effects are confounded; a matched general-purpose Gemini baseline would be needed to isolate embodiment specialization. Raw video is currently available only for the smaller wellness-home subset, so the primary evaluation cannot fully assess temporal visual cues. Future work should therefore run Robotics-ER uniformly across all three SHREC subsets to test cross-context generalization and the proposed externally-grounded--to--internally-inferred performance gradient, and extend the video evaluation to the full wellness-home subset and any additional subsets for which video becomes available, enabling direct comparison with SHREC's video-based baselines. At the team level, future work should test whether independent robots produce compatible social assessments and whether disagreement can be resolved or exploited, then couple those assessments to HVBTA-style closed-loop task allocation with conformal or constraint-based safeguards for high-cost decisions. We additionally evaluate only one embodiment-specialized model family, and SHREC's social labels retain inter-annotator disagreement despite the benchmark's aggregation procedure.

Overall, our preliminary results show that embodiment specialization does not uniformly improve or degrade social reasoning. Gemini Robotics-ER performs better than the general-purpose comparison model on the matched detection condition but worse on social-attribute identification, while additional visual information affects the two tasks differently. This task-dependent pattern suggests that social reasoning should be evaluated explicitly when selecting foundation models for human-aware robot teams.

\bibliographystyle{ieeetr}
\bibliography{mypapers,references,foundationmodels}

@article{lee2025social,
  title={Social Human Robot Embodied Conversation (SHREC) Dataset: Benchmarking Foundational Models' Social Reasoning},
  author={Lee, Dong Won and Kim, Yubin and Guvenoz, Denison and Jeong, Sooyeon and Malachowsky, Parker and Morency, Louis-Philippe and Breazeal, Cynthia and Park, Hae Won},
  journal={arXiv preprint arXiv:2504.13898},
  year={2025}
}

@article{shi2025hribench,
  title={Hribench: Benchmarking vision-language models for real-time human perception in human-robot interaction},
  author={Shi, Zhonghao and Zhao, Enyu and Dennler, Nathaniel and Wang, Jingzhen and Xu, Xinyang and Shrestha, Kaleen and Fu, Mengxue and Seita, Daniel and Matari{\'c}, Maja},
  journal={arXiv preprint arXiv:2506.20566},
  year={2025}
}

@article{liu2026hribench,
  title={HRIBench: Benchmarking Interaction-Centric Human-Robot Collaboration},
  author={Liu, Chang and Zhang, Jiawei and Zhang, Tao and Wang, Ye and Zhou, Hongyu and Jin, Qin},
  journal={arXiv preprint arXiv:2607.13056},
  year={2026}
}

@article{bu2025using,
  title={Using Vision-Language Models as Proxies for Social Intelligence in Human-Robot Interaction},
  author={Bu, Fanjun and Tsai, Melina and Tjokro, Audrey and Bhattacharjee, Tapomayukh and Ortiz, Jorge and Ju, Wendy},
  journal={arXiv preprint arXiv:2512.07177},
  year={2025}
}

@inproceedings{sasabuchi2025agreeing,
  title={Agreeing to interact in human-robot interaction using large language models and vision language models},
  author={Sasabuchi, Kazuhiro and Wake, Naoki and Kanehira, Atsushi and Takamatsu, Jun and Ikeuchi, Katsushi},
  booktitle={2025 34th IEEE International Conference on Robot and Human Interactive Communication (RO-MAN)},
  pages={931--938},
  year={2025},
  organization={IEEE}
}

@article{yeke2026yes,
  title={The Yes-Man Syndrome: Benchmarking Abstention in Embodied Robotic Agents},
  author={Yeke, Doguhan and Temirel, Elif Su and Shreekumar, Ananth and Lee, Brandon and Xu, Dongyan and Celik, Z Berkay},
  journal={arXiv preprint arXiv:2605.20544},
  year={2026}
}

@article{nichols2026should,
  title={How Should I Pick a Foundation Model for My Robot? In Favor of a Community Evaluation Framework for Social Robots},
  author={Nichols, Eric and Markelius, Alva and Gunes, Hatice},
  journal={arXiv preprint arXiv:2608.06898},
  year={2026}
}

@inproceedings{mandi2024roco,
  title={Roco: Dialectic multi-robot collaboration with large language models},
  author={Mandi, Zhao and Jain, Shreeya and Song, Shuran},
  booktitle={2024 IEEE International Conference on Robotics and Automation (ICRA)},
  pages={286--299},
  year={2024},
  organization={IEEE}
}

@inproceedings{kannan2024smart,
  title={Smart-llm: Smart multi-agent robot task planning using large language models},
  author={Kannan, Shyam Sundar and Venkatesh, Vishnunandan LN and Min, Byung-Cheol},
  booktitle={2024 IEEE/RSJ International Conference on Intelligent Robots and Systems (IROS)},
  pages={12140--12147},
  year={2024},
  organization={IEEE}
}

@inproceedings{liu2025coherent,
  title={Coherent: Collaboration of heterogeneous multi-robot system with large language models},
  author={Liu, Kehui and Tang, Zixin and Wang, Dong and Wang, Zhigang and Li, Xuelong and Zhao, Bin},
  booktitle={2025 IEEE International Conference on Robotics and Automation (ICRA)},
  pages={10208--10214},
  year={2025},
  organization={IEEE}
}

@article{wang2024probabilistically,
  title={Probabilistically correct language-based multi-robot planning using conformal prediction},
  author={Wang, Jun and He, Guocheng and Kantaros, Yiannis},
  journal={IEEE Robotics and Automation Letters},
  volume={10},
  number={1},
  pages={160--167},
  year={2024},
  publisher={IEEE}
}

@article{liang2024introspective,
  title={Introspective planning: Aligning robots' uncertainty with inherent task ambiguity},
  author={Liang, Kaiqu and Zhang, Zixu and Fisac, Jaime F},
  journal={Advances in Neural Information Processing Systems},
  volume={37},
  pages={71998--72031},
  year={2024}
}

@article{luo2024sample,
  title={Sample-efficient safety assurances using conformal prediction},
  author={Luo, Rachel and Zhao, Shengjia and Kuck, Jonathan and Ivanovic, Boris and Savarese, Silvio and Schmerling, Edward and Pavone, Marco},
  journal={The International Journal of Robotics Research},
  volume={43},
  number={9},
  pages={1409--1424},
  year={2024},
  publisher={SAGE Publications Sage UK: London, England}
}

@inproceedings{weiner2025hybrid,
  title={Hybrid Voting-Based Task Assignment in Role-Playing Games},
  author={Weiner, Daniel and Korpan, Raj},
  booktitle={Proceedings of the Dungeons, Neurons, and Dialogues: Social Interaction Dynamics in Contextual Games Workshop at HRI'25},
  year={2025}
}

@inproceedings{weiner2025hybridvoting,
  title={Hybrid Voting-Based Task Assignment in Modular Construction Scenarios},
  author={Weiner, Daniel and Korpan, Raj},
  booktitle={Proceedings of the Block by Block: Collaborative Strategies for Multi-agent Robotic Construction Workshop at ICRA'25},
  year={2025}
}

@InProceedings{ichter2022saycan,
  title = 	 {Do As I Can, Not As I Say: Grounding Language in Robotic Affordances},
  author =       {Ichter, Brian and Brohan, Anthony and Chebotar, Yevgen and Finn, Chelsea and Hausman, Karol and Herzog, Alexander and Ho, Daniel and Ibarz, Julian and Irpan, Alex and Jang, Eric and Julian, Ryan and Kalashnikov, Dmitry and Levine, Sergey and Lu, Yao and Parada, Carolina and Rao, Kanishka and Sermanet, Pierre and Toshev, Alexander T and Vanhoucke, Vincent and Xia, Fei and Xiao, Ted and Xu, Peng and Yan, Mengyuan and Brown, Noah and Ahn, Michael and Cortes, Omar and Sievers, Nicolas and Tan, Clayton and Xu, Sichun and Reyes, Diego and Rettinghouse, Jarek and Quiambao, Jornell and Pastor, Peter and Luu, Linda and Lee, Kuang-Huei and Kuang, Yuheng and Jesmonth, Sally and Joshi, Nikhil J. and Jeffrey, Kyle and Ruano, Rosario Jauregui and Hsu, Jasmine and Gopalakrishnan, Keerthana and David, Byron and Zeng, Andy and Fu, Chuyuan Kelly},
  booktitle = 	 {Proceedings of The 6th Conference on Robot Learning},
  pages = 	 {287--318},
  year = 	 {2023},
  editor = 	 {Liu, Karen and Kulic, Dana and Ichnowski, Jeff},
  volume = 	 {205},
  series = 	 {Proceedings of Machine Learning Research},
  month = 	 {14--18 Dec},
  publisher =    {PMLR},
  url = 	 {https://proceedings.mlr.press/v205/ichter23a.html}
}

@InProceedings{zitkovich2023RT2,
  title = 	 {RT-2: Vision-Language-Action Models Transfer Web Knowledge to Robotic Control},
  author =       {Zitkovich, Brianna and Yu, Tianhe and Xu, Sichun and Xu, Peng and Xiao, Ted and Xia, Fei and Wu, Jialin and Wohlhart, Paul and Welker, Stefan and Wahid, Ayzaan and Vuong, Quan and Vanhoucke, Vincent and Tran, Huong and Soricut, Radu and Singh, Anikait and Singh, Jaspiar and Sermanet, Pierre and Sanketi, Pannag R. and Salazar, Grecia and Ryoo, Michael S. and Reymann, Krista and Rao, Kanishka and Pertsch, Karl and Mordatch, Igor and Michalewski, Henryk and Lu, Yao and Levine, Sergey and Lee, Lisa and Lee, Tsang-Wei Edward and Leal, Isabel and Kuang, Yuheng and Kalashnikov, Dmitry and Julian, Ryan and Joshi, Nikhil J. and Irpan, Alex and Ichter, Brian and Hsu, Jasmine and Herzog, Alexander and Hausman, Karol and Gopalakrishnan, Keerthana and Fu, Chuyuan and Florence, Pete and Finn, Chelsea and Dubey, Kumar Avinava and Driess, Danny and Ding, Tianli and Choromanski, Krzysztof Marcin and Chen, Xi and Chebotar, Yevgen and Carbajal, Justice and Brown, Noah and Brohan, Anthony and Arenas, Montserrat Gonzalez and Han, Kehang},
  booktitle = 	 {Proceedings of The 7th Conference on Robot Learning},
  pages = 	 {2165--2183},
  year = 	 {2023},
  editor = 	 {Tan, Jie and Toussaint, Marc and Darvish, Kourosh},
  volume = 	 {229},
  series = 	 {Proceedings of Machine Learning Research},
  month = 	 {06--09 Nov},
  publisher =    {PMLR},
  url = 	 {https://proceedings.mlr.press/v229/zitkovich23a.html}
}

@misc{geminiroboticsteam2025geminiroboticsbringingai,
      title={Gemini Robotics: Bringing AI into the Physical World}, 
      author={Gemini Robotics Team and Saminda Abeyruwan and Joshua Ainslie and Jean-Baptiste Alayrac and Montserrat Gonzalez Arenas and Travis Armstrong and Ashwin Balakrishna and Robert Baruch and Maria Bauza and Michiel Blokzijl and Steven Bohez and Konstantinos Bousmalis and Anthony Brohan and Thomas Buschmann and Arunkumar Byravan and Serkan Cabi and Ken Caluwaerts and Federico Casarini and Oscar Chang and Jose Enrique Chen and Xi Chen and Hao-Tien Lewis Chiang and Krzysztof Choromanski and David D'Ambrosio and Sudeep Dasari and Todor Davchev and Coline Devin and Norman Di Palo and Tianli Ding and Adil Dostmohamed and Danny Driess and Yilun Du and Debidatta Dwibedi and Michael Elabd and Claudio Fantacci and Cody Fong and Erik Frey and Chuyuan Fu and Marissa Giustina and Keerthana Gopalakrishnan and Laura Graesser and Leonard Hasenclever and Nicolas Heess and Brandon Hernaez and Alexander Herzog and R. Alex Hofer and Jan Humplik and Atil Iscen and Mithun George Jacob and Deepali Jain and Ryan Julian and Dmitry Kalashnikov and M. Emre Karagozler and Stefani Karp and Chase Kew and Jerad Kirkland and Sean Kirmani and Yuheng Kuang and Thomas Lampe and Antoine Laurens and Isabel Leal and Alex X. Lee and Tsang-Wei Edward Lee and Jacky Liang and Yixin Lin and Sharath Maddineni and Anirudha Majumdar and Assaf Hurwitz Michaely and Robert Moreno and Michael Neunert and Francesco Nori and Carolina Parada and Emilio Parisotto and Peter Pastor and Acorn Pooley and Kanishka Rao and Krista Reymann and Dorsa Sadigh and Stefano Saliceti and Pannag Sanketi and Pierre Sermanet and Dhruv Shah and Mohit Sharma and Kathryn Shea and Charles Shu and Vikas Sindhwani and Sumeet Singh and Radu Soricut and Jost Tobias Springenberg and Rachel Sterneck and Razvan Surdulescu and Jie Tan and Jonathan Tompson and Vincent Vanhoucke and Jake Varley and Grace Vesom and Giulia Vezzani and Oriol Vinyals and Ayzaan Wahid and Stefan Welker and Paul Wohlhart and Fei Xia and Ted Xiao and Annie Xie and Jinyu Xie and Peng Xu and Sichun Xu and Ying Xu and Zhuo Xu and Yuxiang Yang and Rui Yao and Sergey Yaroshenko and Wenhao Yu and Wentao Yuan and Jingwei Zhang and Tingnan Zhang and Allan Zhou and Yuxiang Zhou},
      year={2025},
      eprint={2503.20020},
      archivePrefix={arXiv},
      primaryClass={cs.RO},
      url={https://arxiv.org/abs/2503.20020}, 
}

@misc{lee2026shrec,
      title={Social Human Robot Embodied Conversation (SHREC) Dataset: Benchmarking Foundational Models' Social Reasoning}, 
      author={Dong Won Lee and Yubin Kim and Denison Guvenoz and Sooyeon Jeong and Parker Malachowsky and Louis-Philippe Morency and Cynthia Breazeal and Hae Won Park},
      year={2026},
      eprint={2504.13898},
      archivePrefix={arXiv},
      primaryClass={cs.HC},
      url={https://arxiv.org/abs/2504.13898}, 
}

\end{document}